\documentclass[dvipdfmxm, pdflatex]{article}
\usepackage{spconf,amsmath,graphicx}

\usepackage{epsfig}
\usepackage{multirow}
\usepackage{setspace}
\usepackage{booktabs}
\usepackage{cleveref}
\usepackage[hyphens]{url}
\usepackage[subrefformat=parens]{subcaption}
\usepackage{color}
\usepackage{amssymb}

\makeatletter
\def\eg{\emph{e.g}.}

\makeatother

\crefname{figure}{Fig.}{Fig.}
\crefname{table}{Table}{Table}

\newcommand\beginsupplement{%
        \setcounter{table}{0}
        \renewcommand{\thetable}{\Alph{table}}%
        \setcounter{figure}{0}
        \renewcommand{\thefigure}{\Alph{figure}}%
     }

\newcommand{\figMainTeaserImageNetX}{
    \begin{figure}[t]
      \centering
      \centerline{\includegraphics[width=\linewidth]{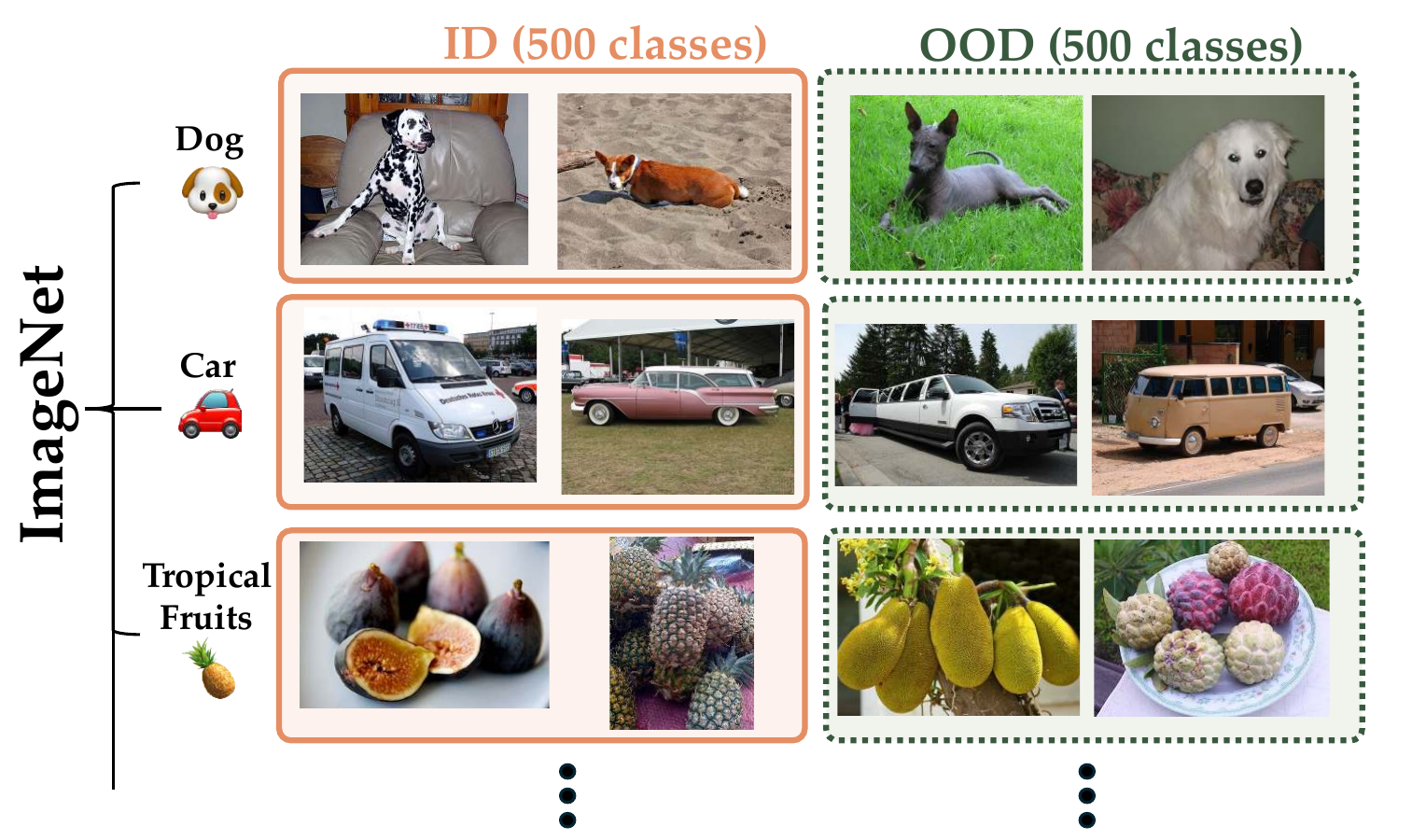}}
    \caption{Data examples of ID and OOD in ImageNet-X, a benchmark for challenging semantic shifts.}
    \label{fig:teaser_imagenet_x}
    \end{figure}
}

\newcommand{\figMainTeaserBigVer}{
\begin{figure*}[t]
\begin{tabular}{ccc}
       \begin{minipage}[t]{0.33\hsize}
        \centering
        \includegraphics[keepaspectratio, scale=0.24]{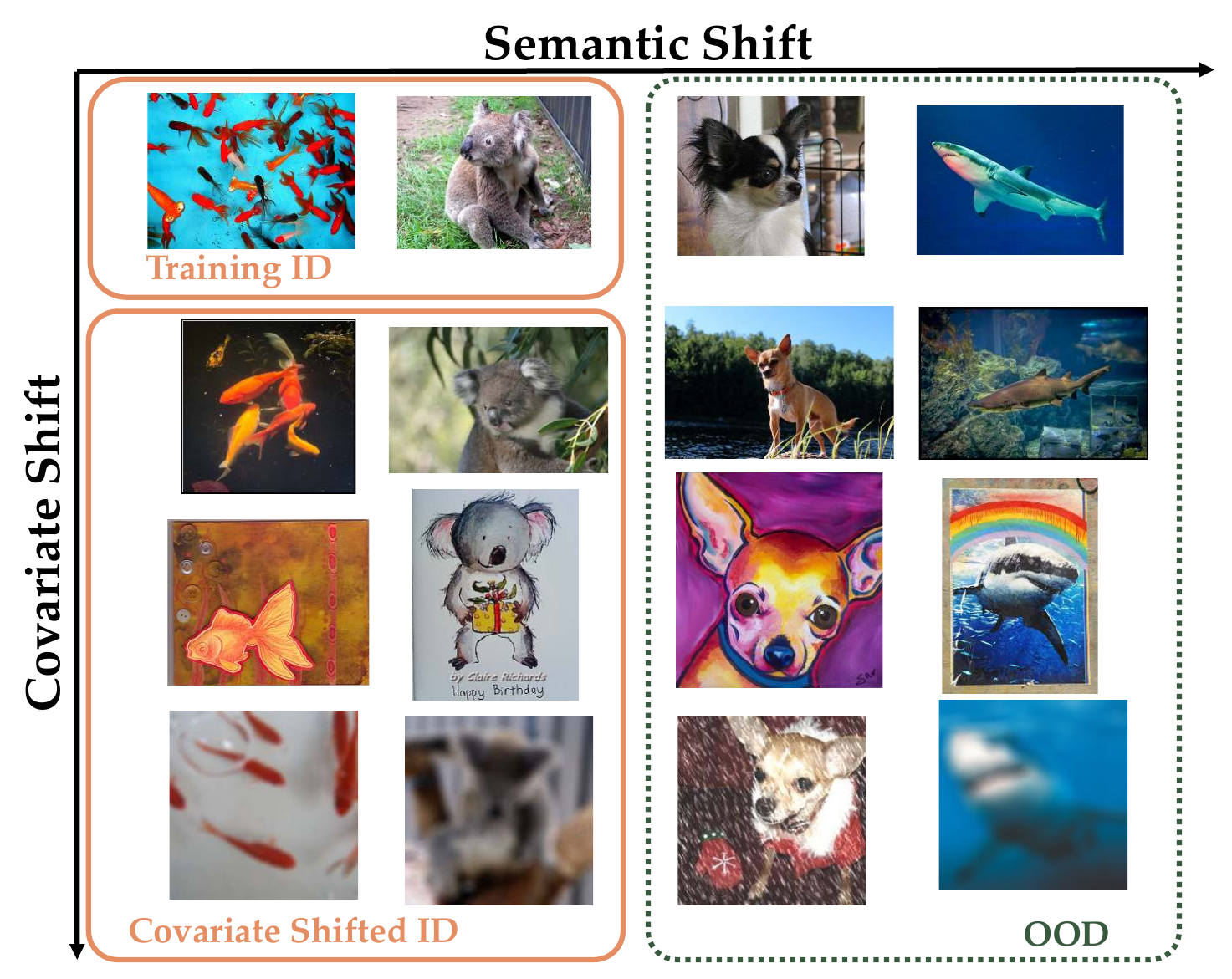}\\
        \captionsetup{justification=centering}
        \subcaption{ImageNet-FS-X.}
        \label{fig:imagenet_fs_x_benchmark}
      \end{minipage} 
      \begin{minipage}[t]{0.33\hsize}
        \centering
        \includegraphics[keepaspectratio, scale=0.24]{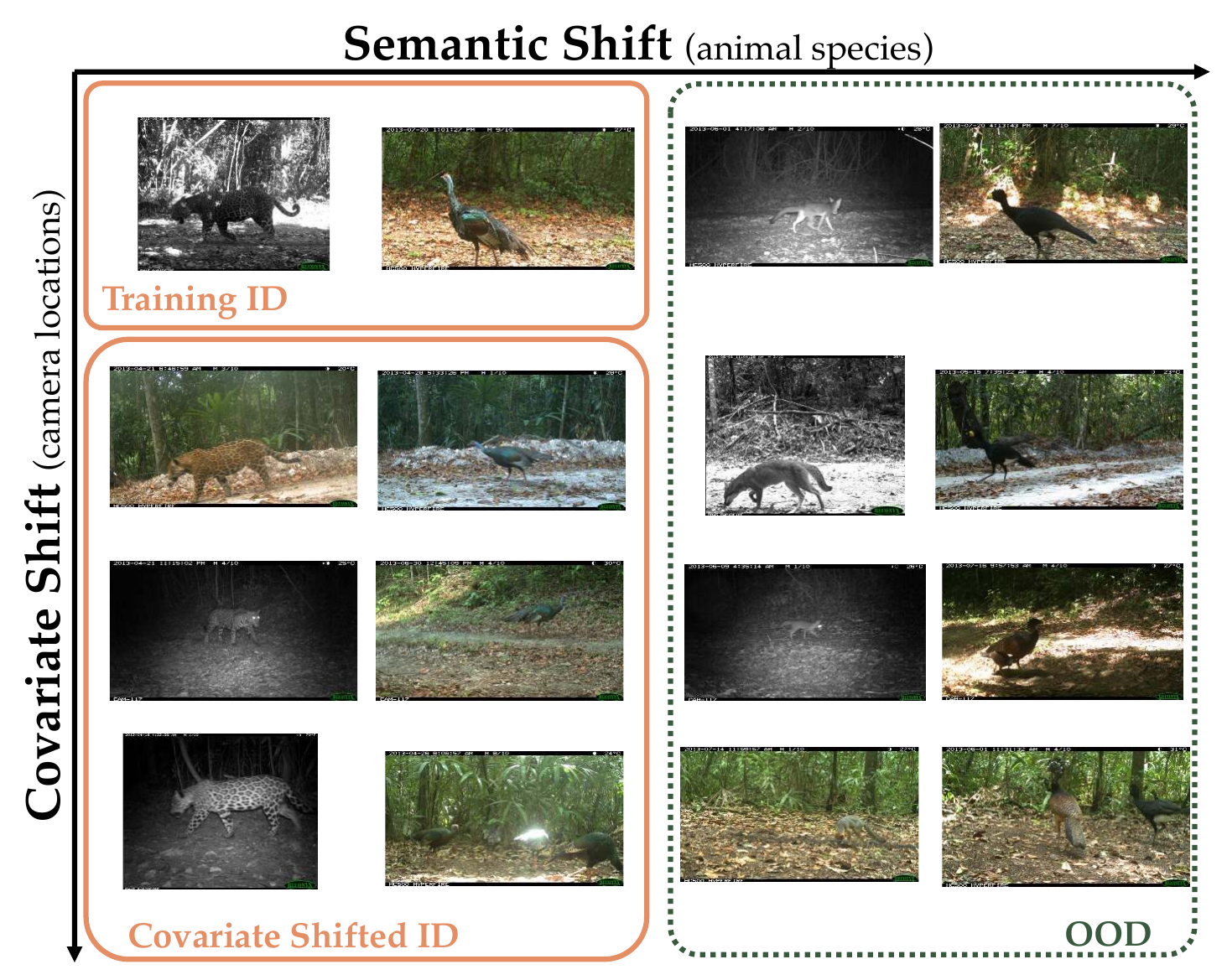}\\
        \captionsetup{justification=centering}
        \subcaption{Wilds-FS-X (iWildCam).}
        \label{fig:iwildcam_fs_x_benchmark}
      \end{minipage}
      \begin{minipage}[t]{0.33\hsize}
        \centering
        \includegraphics[keepaspectratio, scale=0.24]{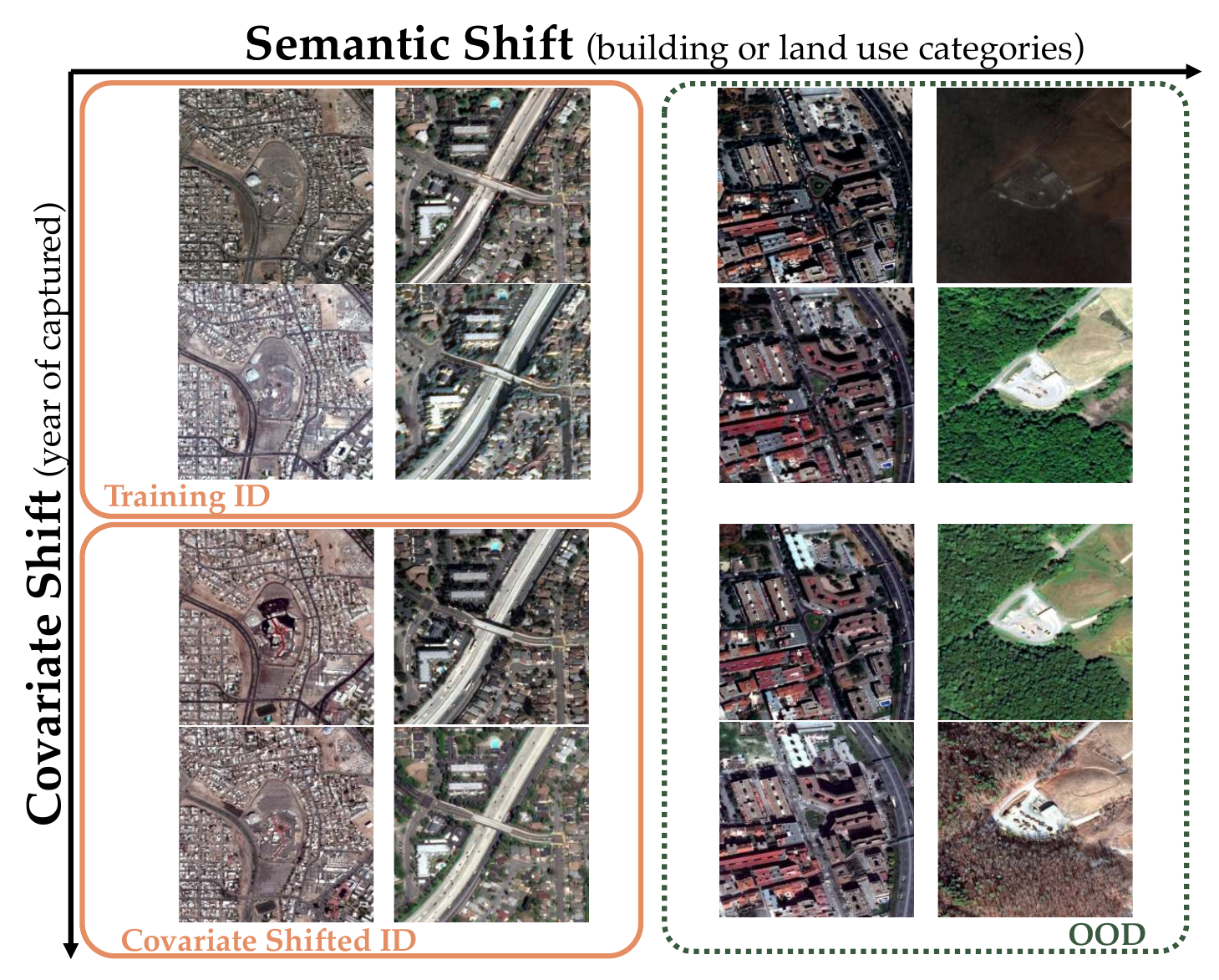}\\
        \captionsetup{justification=centering}
        \subcaption{Wilds-FS-X (FMoW).}
        \label{fig:fmow_fs_x_benchmark}
      \end{minipage}\\
  \end{tabular}
    \caption{Data structure and examples in the proposed benchmarks incorporating covariate shifts (feature distribution shifts).}
    \label{fig:teaser_big}
\end{figure*}
}

\newcommand{\tabDataset}{
\begin{table}[t]
    \footnotesize
    \begin{spacing}{0.6}
     \caption{Data composition of each benchmark. \\The number of classes and samples (used in this experiment) assigned to ID and OOD for each source dataset are shown. }
           \label{tab:dataset_data}
     \end{spacing}
     \vskip 5pt
     \centering
      \begin{tabular}{@{}ccrrrr@{}}
       \toprule
       &&\multicolumn{2}{c}{\textbf{\#Classes}}&\multicolumn{2}{c}{\textbf{\#Samples}}\\
        \cmidrule(lr){3-4} \cmidrule(lr){5-6}
       &\textbf{Source}&ID&OOD&ID&OOD\\
       \midrule
       ImageNet-X&ImageNet-1k&500&500&25,000&25,000\\
       \midrule
       \multirow{3}{*}{\shortstack{ImageNet-\\FS-X}}&ImageNet-V2&500&500&5,000&5,000\\
        &ImageNet-R&495&492&14,755&15,245\\
       &ImageNet-C&315&313&5,380&4,620\\
       \midrule
       \multirow{4}{*}{\shortstack{Wilds-FS-X}}&iWildCam \textit{Test-ID}&91&91&3,801&4,353\\
       &iWildCam \textit{Test}&91&91&14,526&28,265\\
       &FMoW \textit{Test-ID}&31&31&5,109&6,218\\
       &FMoW \textit{Test}&31&31&9,601&12,507\\
       \bottomrule
  \end{tabular}
\end{table}
}

\newcommand{\tabWildsAccuracy}{
\begin{table}[t]
    \small
    \begin{spacing}{0.6}
     \caption{Classification accuracy (\%) of representative CLIP-based methods on Wilds-FS-X.}
           \label{tab:compare_wilds_accuracy}
     \end{spacing}
     \vskip 5pt
     \centering
      \begin{tabular}{@{}ccllll@{}}
       \toprule
       &&\multicolumn{2}{c}{iWildCam}&\multicolumn{2}{c}{FMoW}\\
        \cmidrule(lr){3-4} \cmidrule(lr){5-6}
       \multicolumn{2}{c}{methods}&Test-ID&Test&Test-ID&Test\\
       \midrule
       \textit{Zero-Shot}&&7.63&14.68&20.14&22.56\\
       \midrule
       \multirow{4}{*}{\textit{Few-Shot}}&CoOp~\cite{CoOp_zhou2022learning}&\underline{40.79}&\underline{41.44}&\underline{37.97}&\underline{39.53}\\
       &LoCoOp~\cite{LoCoOp_miyai2024locoop}&\textbf{45.43}&\textbf{43.10}&\textbf{38.57}&\textbf{40.25}\\
       &NegPrompt~\cite{NegativePrompts_li2024learning}&36.24&37.45&35.62&37.89\\
       &IDPrompt~\cite{IDLike_bai2024id}&9.27&12.38&24.94&27.64\\
       \bottomrule
  \end{tabular}
\end{table}
}

\newcommand{\figImageNetXFullSpectrum}{
\begin{figure}[t]
  \centering
  \centerline{\includegraphics[width=\linewidth]{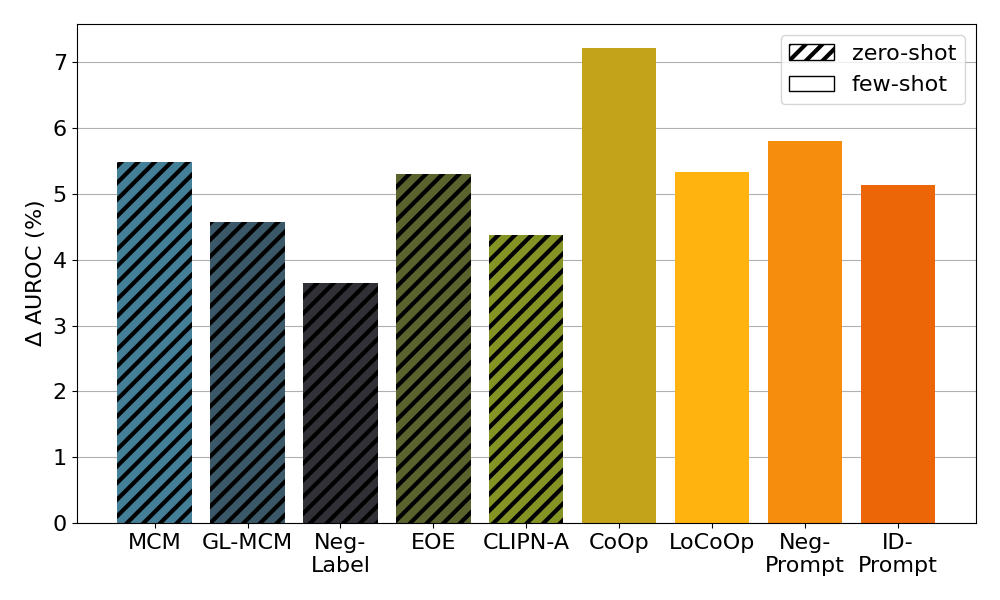}}
\caption{Difference in OOD detection performance (AUROC) between ImageNet-X and ImageNet-FS-X for CLIP-based methods, calculated as ImageNet-X minus ImageNet-FS-X results.}
\label{fig:compare_sa_imagenet}
\end{figure}
}

\newcommand{\tabCombineOODAll}{
\begin{table*}[t]
    \small
    \begin{spacing}{0.6}
     \caption{Comparison of OOD detection performance of representative CLIP-based methods (AUROC$\uparrow$). \\ The performance rankings changed across the benchmarks.
No single winner emerges as the best across all benchmarks. }
           \label{tab:compare_ood_result}
     \end{spacing}
     \vskip 5pt
     \centering
     \begin{tabular}{@{}ccllllll@{}}
       \toprule
       &&\multicolumn{2}{c}{Conventional Benchmark}&ImageNet-X&ImageNet-FS-X&\multicolumn{2}{c}{Wilds-FS-X}\\
        \cmidrule(lr){3-4} \cmidrule(lr){7-8}
       \multicolumn{2}{c}{methods}&Common-OOD&Hard-OOD&&&iWildCam&FMoW\\
       \midrule
       \multirow{5}{*}{\textit{Zero-Shot}}&MCM\cite{MCM_ming2022delving}&90.52 &63.72 &74.32 & 68.83 &\underline{76.42} &52.80 \\
       &GL-MCM\cite{GLMCM_miyai2023zero}&92.08 &67.43 &75.36 & 70.79&72.46 &52.85\\
       &NegLabel~\cite{NegPrompt_jiangnegative}&\textbf{94.83} &70.44&72.15 & 68.50&\textbf{78.74}&53.73\\
       &EOE~\cite{EOE_caoenvisioning}&93.11 &67.85 &77.07 &\underline{71.76} & 62.71&55.45\\
       &CLIPN-A~\cite{CLIPN_wang2023clipn}&\underline{93.76} &\textbf{82.51} &\underline{77.08} & 69.81&61.60&55.89\\
       \midrule
       \multirow{4}{*}{\textit{Few-Shot}}&CoOp~\cite{CoOp_zhou2022learning}&92.38 &67.54 &76.63 & 69.41&71.35&\textbf{56.44}\\
       &LoCoOp~\cite{LoCoOp_miyai2024locoop}&93.49 & 67.77 &73.74 &68.40 &65.50&55.99\\
       &NegPrompt~\cite{NegativePrompts_li2024learning}&92.73 &\underline{76.91}&\textbf{80.95} &\textbf{75.14}&57.62&\underline{56.33}\\
       &IDPrompt~\cite{IDLike_bai2024id}&90.35 &60.43 &67.42 & 62.28&60.18 &51.37 \\
       \bottomrule
  \end{tabular}
\end{table*}
}
\newcommand{\figImageNetFSCompare}{
    \begin{figure*}[t]
      \centering
      \centerline{\includegraphics[width=\linewidth]{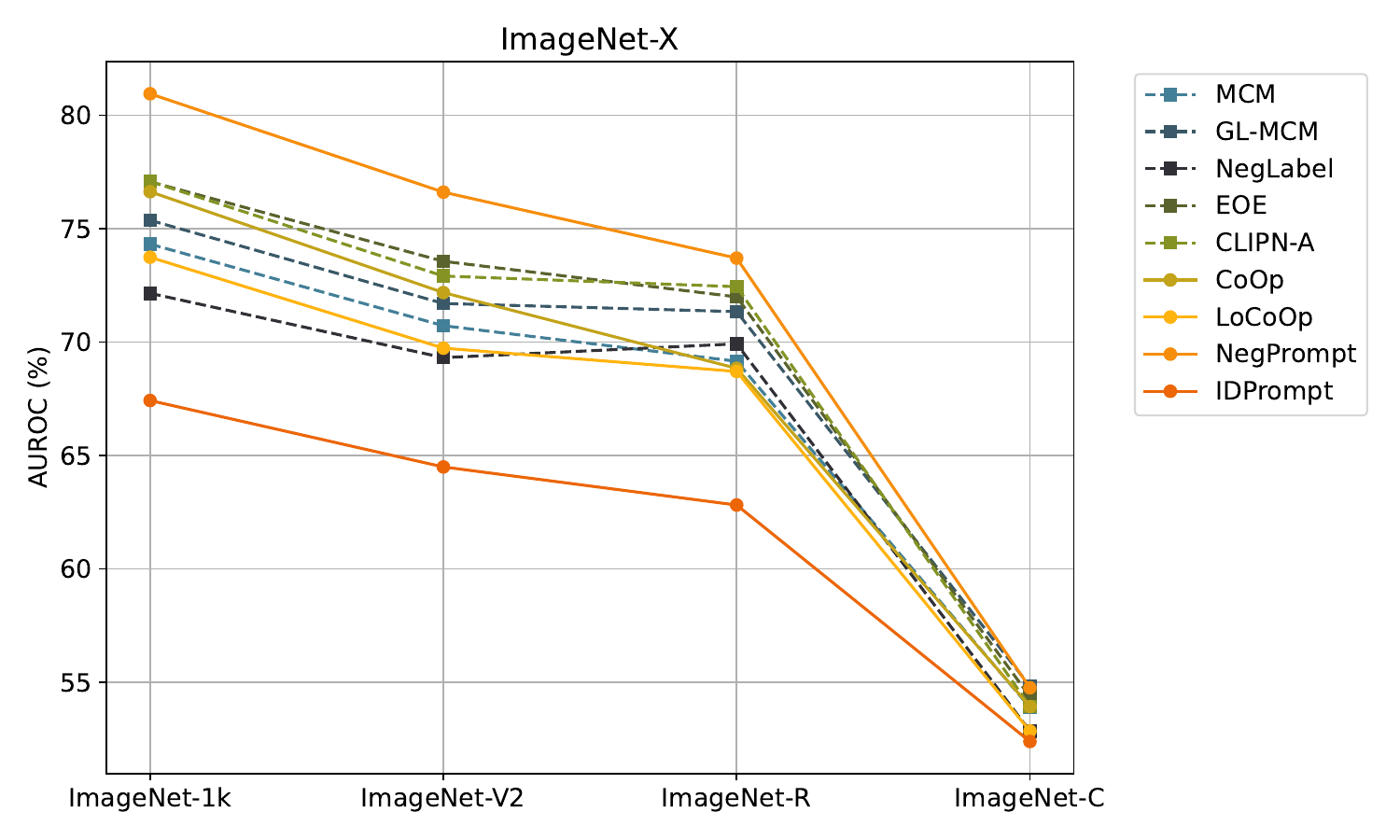}}
    \caption{OOD detection performance for each source dataset in ImageNet-FS-X (AUROC$\uparrow$). Dashed lines represent zero-shot methods, while solid lines indicate few-shot methods.}
    \label{fig:compare_imagenet_fs}
    \end{figure*}
}

\newcommand{\figWildsFSCompare}{
    \begin{figure*}[t]
      \centering
      \centerline{\includegraphics[width=\linewidth]{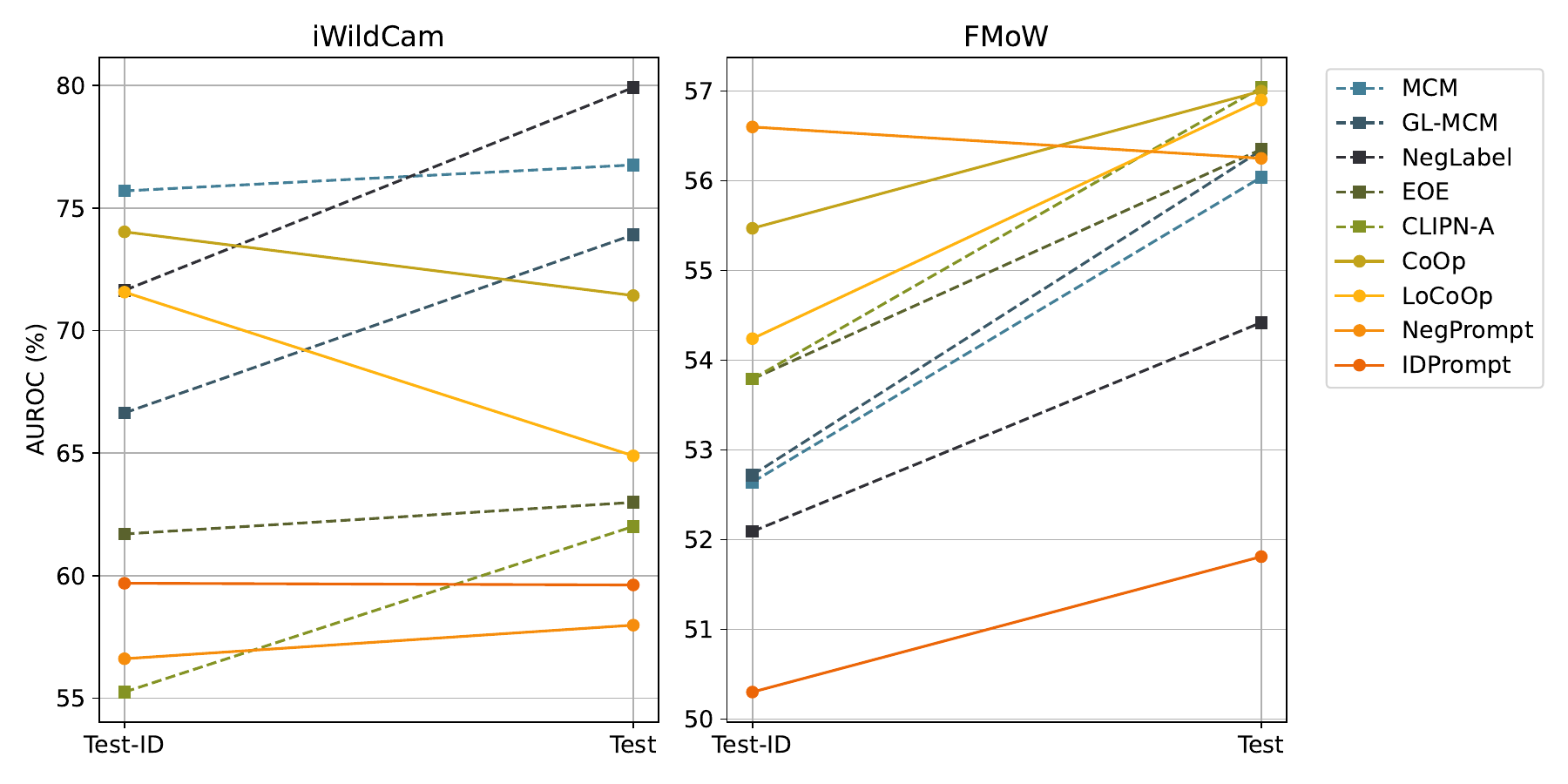}}
    \caption{OOD detection performance for each source dataset in Wilds-FS-X (AUROC$\uparrow$). Dashed lines represent zero-shot methods, while solid lines indicate few-shot methods.}
    \label{fig:compare_wilds_fs}
    \end{figure*}
}

\newcommand{\figImageNetFSXWithSample}{
    \begin{figure*}[t]
      \centering
      \centerline{\includegraphics[width=\linewidth]{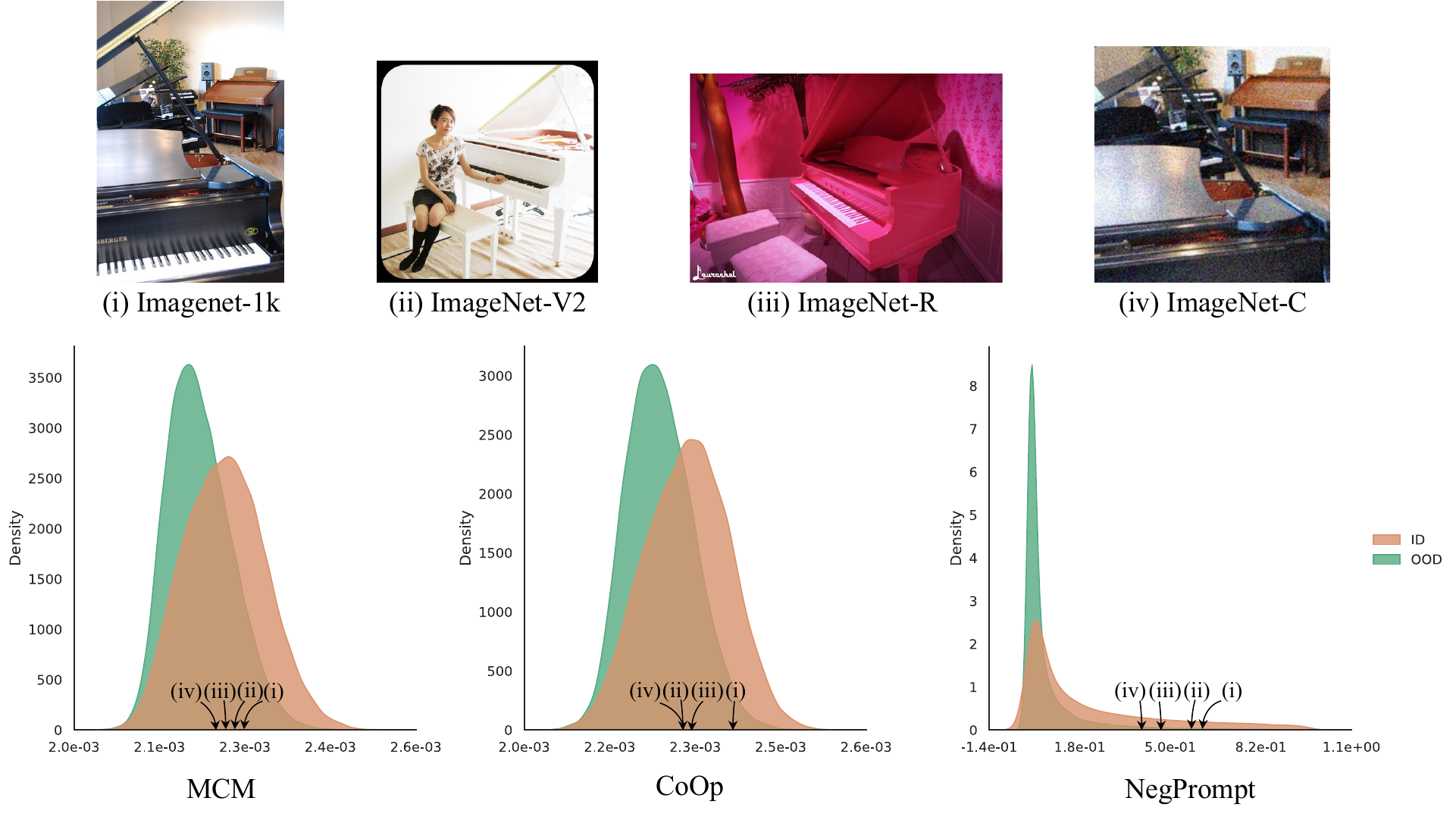}}
    \caption{OOD score distribution and sample scores in ImageNet-FS-X. The samples belong to the ``piano'' label and are part of the ID data. }
    \label{fig:distribution_imagenet_fs_x_ver2}
    \end{figure*}
}

\newcommand{\figImageNetXWithSample}{
    \begin{figure*}[t]
      \centering
      \centerline{\includegraphics[width=\linewidth]{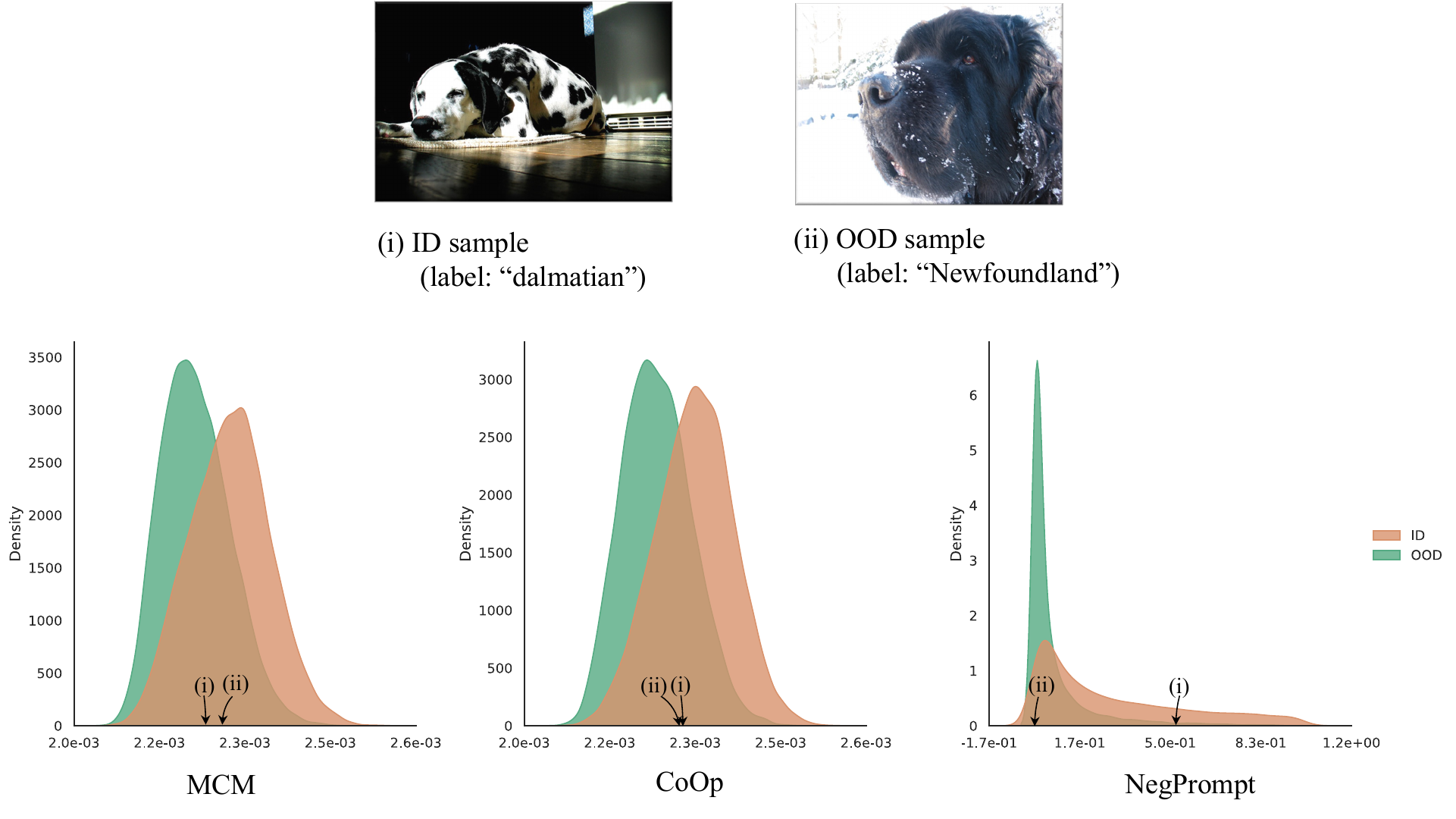}}
    \caption{OOD score distribution and sample scores in ImageNet-X.}
    \label{fig:distribution_imagenet_x_ver2}
    \end{figure*}
}

\title{A BENCHMARK AND EVALUATION FOR REAL-WORLD OUT-OF-DISTRIBUTION DETECTION USING VISION-LANGUAGE MODELS}
\name{Shiho Noda$^{\star}$, Atsuyuki Miyai$^{\star}$, Qing Yu$^{\star}$, Go Irie$^{\dagger}$, Kiyoharu Aizawa$^{\star \dagger}$}
\address{The University of Tokyo, Japan$^{\star}$\\
        Tokyo University of Science, Japan$^{\dagger}$}
\begin{document}

% \ninept
%
\maketitle
%
% 100~150 word
\begin{abstract}

Out-of-distribution (OOD) detection is a task that detects OOD samples during inference to ensure the safety of deployed models. However, conventional benchmarks have reached performance saturation, making it difficult to compare recent OOD detection methods. To address this challenge, we introduce three novel OOD detection benchmarks that enable a deeper understanding of method characteristics and reflect real-world conditions. First, we present ImageNet-X, designed to evaluate performance under challenging semantic shifts. Second, we propose ImageNet-FS-X for full-spectrum OOD detection, assessing robustness to covariate shifts (feature distribution shifts). Finally, we propose Wilds-FS-X, which extends these evaluations to real-world datasets, offering a more comprehensive testbed. Our experiments reveal that recent CLIP-based OOD detection methods struggle to varying degrees across the three proposed benchmarks, and none of them consistently outperforms the others. We hope the community goes beyond specific benchmarks and includes more challenging conditions reflecting real-world scenarios. 
The code is \url{https://github.com/hoshi23/OOD-X-Benchmarks}.

\end{abstract}

\section{Introduction}
\vspace{-7pt}
\label{sec:intro}

Machine learning models, including vision-language models (VLMs), have recently improved image recognition but may fail with unknown examples, risking real-world applications~\cite{AIkiken_amodei2016concrete, AIkiken2_mohseni2022taxonomy}.
% In recent years, machine vision models, including vision-language models (VLMs), have significantly improved image recognition accuracy.
% However, these models may produce incorrect results when they encounter unknown examples.
% This poses a substantial risk in real-world automation applications~\cite{AIkiken_amodei2016concrete, AIkiken2_mohseni2022taxonomy}.
Out-of-distribution (OOD) detection, which identifies samples that deviate from the semantics distribution of the in-distribution (ID) data, is crucial for ensuring model reliability~\cite{OODSurvey_yang2024generalized}.
% OOD detection aims to identify OOD samples that deviate from the semantics distribution of the in-distribution (ID) data~\cite{OODSurvey_yang2024generalized}.
Recently, VLM-based methods like CLIP~\cite{CLIP_radford2021learning},  have attracted attention, inspiring many approaches in this field~\cite{MCM_ming2022delving, LoCoOp_miyai2024locoop}.

Various methods have been proposed, and high OOD detection performance has been achieved.
However, conventional benchmarks have reached performance saturation, making it difficult to distinguish between different approaches.
Commonly used evaluation benchmarks include using ImageNet-1k~\cite{ImageNet1k_russakovsky2015imagenet} as ID data and other datasets like iNaturalist~\cite{iNaturalist_van2018inaturalist} or Texture~\cite{Texture_cimpoi2014describing} as OOD data. 
These benchmarks have served as widely adopted baselines for comparing the OOD detection performance~\cite{MCM_ming2022delving, LoCoOp_miyai2024locoop,GLMCM_miyai2023zero, NegPrompt_jiangnegative,EOE_caoenvisioning,CLIPN_wang2023clipn,
 NegativePrompts_li2024learning, IDLike_bai2024id}.
However, the large distribution shift between ID and OOD data makes detection easier, making it difficult to analyze and compare the characteristics of different methods.

\figMainTeaserImageNetX
\figMainTeaserBigVer

In this paper, we propose three novel benchmarks, supporting a precise understanding of method characteristics and simulations reflecting real-world conditions (\cref{fig:teaser_imagenet_x}, \cref{fig:teaser_big}).
For challenging and practical evaluations, three critical perspectives are crucial: (1) the semantic similarity between ID and OOD data, (2) two distribution shifts - semantic shift (class shift) and covariate shift (feature distribution shift), and (3) the dataset's alignment with real-world scenarios. 
Here, we define label shifts as `semantic shifts,' and changes in image style or quality as `covariate shifts' following~\cite{FullSpec_yang2023full}.
Research addressing both as mentioned in (2) is termed full-spectrum OOD detection~\cite{FullSpec_yang2023full}.
These perspectives reflect real-world scenarios and enable us to conduct more robust evaluations across different OOD detection methods.

Based on the above perspectives, we propose: (B1) \textbf{ImageNet-X}, a benchmark with a small semantic shift, 
It is created by splitting ID and OOD data from ImageNet-1k~\cite{ImageNet1k_russakovsky2015imagenet}, considering the hierarchical structure of ImageNet to ensure that the semantic similarity between ID and OOD data is small.
(B2) \textbf{ImageNet-FS-X}, a benchmark that incorporates covariate shift into (B1), allowing for a systematic analysis of these two distribution shifts. (B3) \textbf{Wilds-FS-X}, a benchmark that retains the characteristics of (B1) and (B2) while bringing the data closer to real-world scenarios. This benchmark is based on the Wilds~\cite{WILDS_wilds2021} and reflects distribution shifts that naturally occur in real-world application scenarios. Evaluating methods using these three progressive datasets enables rigorous analysis under complex, real-world conditions. 

In our experiments, we evaluate representative CLIP-based OOD detection methods.
% While these methods have made significant advancements in recent years, they have not been sufficiently evaluated and compared under unified conditions, especially under conditions of challenging semantic shift and incorporating covariate shift.
% Therefore, we demonstrate the current state of OOD detection with VLMs through evaluation using our benchmarks.
Evaluation using the three proposed benchmarks revealed the following insights:
(B1) Comparing with the results of the conventional benchmarks, a change in the method performance ranking is observed, and no single winner emerges as the best across all distribution shifts.
(B2) CLIP-based OOD detection performance decreases due to covariate shifts, but the method performance ranking remains consistent with constant semantic shift.
(B3) Both classification and OOD detection performance are low, making it a highly challenging benchmark, and even though few-shot improved classification accuracy, it does not necessarily improve OOD detection performance.
% \miyai{データセット論文なのにfindingsがあっさりしている。データセットごとに一番重要なfidingsを書いてみると良いと思う。}

Major contributions of the paper are:

\vspace{-7pt}
\begin{spacing}{0.5}
\begin{itemize}
    \item We proposed three benchmarks for challenging and practical evaluations, enabling a precise understanding of method characteristics and simulations of real-world conditions.
    \item Our experiments revealed performance variations across benchmarks and significant drops when incorporating covariate shifts and real-world scenarios, highlighting the need for improvement.
\end{itemize}
\end{spacing}

% This paper suggests that to advance the OOD detection field, evaluations must go beyond specific benchmarks and incorporate more challenging conditions that reflect real-world scenarios.

\vspace{-10pt}
\section{Benchmark Construction}\vspace{-5pt}
\label{sec:benchmarks}

This section outlines the background and details the construction methods of each of the three proposed benchmarks.

\subsection{ImageNet-X}\vspace{-5pt}
\label{ssec:imagenet_x}
\textbf{Background.}
For OOD detection, conventional benchmarks have primarily evaluated the detection of large distribution shifts, including both semantic and covariate shifts (common OOD detection) with ImageNet~\cite{MSP_hendrycks2022baseline,mos_huang2021mos}.
In contrast, hard OOD detection has recently been proposed as a problem setting focusing on detecting small semantic shifts between labels.
For example, some benchmarks use ImageNet-1k as ID data and use tailored datasets~\cite{SSBHard_vaze2022open, NINCO_bitterwolf2023or} as OOD data~\cite{OpenoodV15_zhang2023openood}.
However, using different datasets for ID and OOD may unintentionally incorporate covariate shifts.
OOD detection performance is more sensitive to covariate shifts than semantic shifts~\cite{ImageNetOOD_yangimagenet}.
Therefore, utilizing both ID and OOD data from a common dataset is considered an effective approach to minimize the impact of covariate shifts, enabling a more rigorous evaluation of semantic shifts.

Additionally, ImageNet-10 and ImageNet-20~\cite{MCM_ming2022delving}, as well as ImageNet-Protocol~\cite{opensetImagenet_palechor2023large}, are benchmarks that use datasets split from ImageNet.
However, the former has a limited number of classes, and the latter suffers from a considerable semantic gap between ID and OOD, resulting in performance saturation for both.
Therefore, a dataset with more classes and guaranteed similarity in semantic shifts is necessary for more accurate performance comparisons.

\noindent\textbf{Proposed Benchmark.}
We propose ImageNet-X for large-scale and more rigorous evaluation of challenging semantic shifts. 
In ImageNet-X, we propose a splitting method for ID and OOD labels based on their semantic structure, dividing closely related labels within that structure.
For example, we treat ``dalmatian'' as ID and ``Great Pyrenees'' as OOD (\cref{fig:teaser_imagenet_x}).
We use the WordNet hierarchy~\cite{wordnet} to represent the labels' semantic structure, with each ImageNet label corresponding to a parent class. 
By using the parent class data, which is the immediate upper-level hierarchy provided on the official ImageNet site~\cite{ImageNet1k_russakovsky2015imagenet}, the 1000 classes can be grouped into 558 parent classes. 
This allows for the separation of labels that are particularly closely related.

The splitting process is conducted as follows:
First, the labels within each parent class are split into two halves.
Next, the ID and OOD labels are balanced by integrating the divided labels, resulting in an equal number of labels for ID and OOD.
The number of split classes and samples are shown in \cref{tab:dataset_data}.

\vspace{-5pt}
\subsection{ImageNet-FS-X}\vspace{-3pt}
\label{ssec:imagenet_fs_x}
\textbf{Background.}
An existing problem setting for OOD detection with covariate shifts is full-spectrum OOD (FS-OOD)~\cite{FullSpec_yang2023full,OpenoodV15_zhang2023openood}.
In FS-OOD, the ID data is constructed by adding data with the same semantics but different covariate distributions (covariate-shifted ID) to the training data (training ID).
When using ImageNet-1k as the training ID, variant datasets (\eg, ImageNet-V2~\cite{Imagenetv2_recht2019imagenet}, ImageNet-R~\cite{ImagenetR_hendrycks2021many}, and ImageNet-C~\cite{ImageNetC_hendrycks2019robustness}) are used as covariate-shifted IDs (ImageNet-FS).
Covariate-shifted ID is added only during evaluation to assess the performance change of ID data due to covariate shift.
However, differences in covariate distributions between ID and OOD data are not considered, making it difficult to systematically evaluate semantic and covariate shifts separately.
Moreover, ~\cite{covariateParadox_long2024rethinking} proposed a benchmark that allows flexible evaluation of semantic shift and covariate shift in OOD data and analyzed their respective impacts. 
However, this does not account for covariate shifts in ID data and does not cover the FS-OOD problem setting.

\noindent\textbf{Proposed Benchmark.} We propose ImageNet-FS-X, which incorporates covariate shift into ImageNet-X (\cref{fig:imagenet_fs_x_benchmark}). This involves adding data with the same labels but different covariate distributions to ImageNet-1k and applying the label splitting method from ImageNet-X to construct ID and OOD data.

The data in ImageNet-FS, specifically the training and variant datasets, are separated into ID and OOD data based on their labels. 
The ID data from the training dataset is used as the training ID, and the ID data from the variant datasets is used as the covariate-shifted ID. 
On the other hand, the other data is collectively used as OOD data.
The number of split classes and samples for each derived dataset are shown in \cref{tab:dataset_data}.
This approach aligns the covariate distribution of OOD data with ID data, enabling a more rigorous evaluation of semantic shift and the impact of covariate shifts on both ID and OOD data.
The ImageNet-FS-X benchmark allows for a comprehensive evaluation that considers both sensitivity to semantic shifts and robustness to covariate shifts.

\tabDataset
\vspace{-8pt}
\subsection{Wilds-FS-X}\vspace{-5pt}
\label{ssec:wilds_fs_x}
\textbf{Background.}
An existing benchmark using real-world scenario datasets is WildCapture~\cite{wildcapture_10350906}, which evaluates using a dataset of wildlife trap images. 
However, it does not explicitly separate the covariate distribution differences between the training and evaluation data, and thus, robustness to covariate shifts has not been evaluated.

\noindent\textbf{Proposed Benchmark.}
Finally, we propose Wilds-FS-X, a benchmark that brings the problem setting of ImageNet-FS-X closer to real-world scenarios by utilizing Wilds~\cite{WILDS_wilds2021}.
Wilds is a benchmark that reflects covariate shifts naturally occurring in real-world scenarios. 
This benchmark evaluates task performance by using test data (\textit{Test}) that shares the same labels as the training data (\textit{Training}) but differs in domains such as shooting environments. 
Wilds aims to evaluate the impact of covariate shifts on recognition performance.
To facilitate performance comparison, a test dataset with identical covariate distributions to the training data (\textit{Test-ID}) is also included.

The model is trained using the ID data from \textit{Training}. 
For evaluation, the ID data from \textit{Test-ID} is defined as training ID, the ID data from \textit{Test} is defined as covariate-shifted ID, and the remaining data from \textit{Test-ID} and \textit{Test} are defined as OOD.

Wilds consists of datasets spanning 10 different domains. 
For this benchmark, we utilized: iWildCam and FMoW.

iWildCam is a dataset of wildlife photos taken by camera traps.
Semantic shift corresponds to different animal species, and covariate shift corresponds to changes in camera locations of camera photos (\cref{fig:iwildcam_fs_x_benchmark}). 
The labels comprise 182 classes, including 181 animal species, and an additional class labeled ``empty'' for images containing no animals. 
Among these, 91 animal labels were randomly selected as ID data, while the remaining 90 animal labels, along with the ``empty'' class, were used as OOD data.
FMoW is a RGB satellite image dataset. 
Semantic shift corresponds to the building or land use categories, and covariate shift corresponds to changes in the year when the images were captured (\cref{fig:fmow_fs_x_benchmark}). 
The dataset includes 62 facility types. 
From these, 31 classes were randomly chosen as ID data, while the remaining 31 classes were designated as OOD data.
The specific sample sizes for the splits in this experiment are shown in \cref{tab:dataset_data}.
Using these datasets, we can evaluate the detection of fine-grained semantic shifts and robustness to covariate shifts required in real-world scenarios.

\vspace{-7pt}
\section{Experiments}\vspace{-5pt}
\label{sec:exmeriments}

\subsection{Baseline Methods}\vspace{-3pt}

We examine representative zero-shot and few-shot CLIP-based OOD detection methods.
Zero-shot and few-shot settings are the most common settings for CLIP-based OOD detection.

\noindent\textbf{Zero-shot OOD detection.}
Zero-shot OOD detection is a method that does not require additional training with ID data~\cite{OODVlmSurvey_miyai2024generalized}. 
Zero-shot methods can be broadly categorized into two types.
The first type does not use an OOD prompt and relies solely on the outputs of each ID class label from CLIP to calculate OOD scores. 
We utilize MCM~\cite{MCM_ming2022delving} and GL-MCM~\cite{GLMCM_miyai2023zero} in our experiments.
The second type utilizes OOD prompts, which involve additional labels or prompts specifically designed for OOD detection. 
We utilize NegLabel~\cite{NegPrompt_jiangnegative}, EOE~\cite{EOE_caoenvisioning}, and CLIPN~\cite{CLIPN_wang2023clipn} in our experiments.

\noindent\textbf{Few-shot OOD detection.}
Few-shot OOD detection refers to the use of ID images during both the training and inference phases~\cite{OODVlmSurvey_miyai2024generalized}
We utilize several methods in which the context of prompts is learned as learnable parameters from few-shot ID data. 
Few-shot methods are also divided into approaches that use OOD prompts and those that do not. 
For methods without OOD prompts, we use CoOp~\cite{CoOp_zhou2022learning} and LoCoOp~\cite{LoCoOp_miyai2024locoop}. 
The hyperparameters are set according to previous studies, with training performed using 16 shots in all cases. 
The OOD score calculation for these methods can apply the same approach as zero-shot methods, such as MCM and GL-MCM.
In this experiment, following the respective papers, CoOp is evaluated using MCM, and LoCoOp is evaluated using GL-MCM. 
For few-shot methods that use OOD prompts, we use NegPrompt~\cite{NegativePrompts_li2024learning} and IDPrompt~\cite{IDLike_bai2024id}. 
These methods involve additional training on OOD prompts.
In NegPrompt, the hyperparameters from prior work are applied, and training is done with 16 shots. 
On the other hand, IDPrompt has a higher learning cost, therefore, following previous studies, training is conducted with just 1 shot. 
The average value for each of the few-shot methods was calculated over 3 seeds.

\vspace{-5pt}
\subsection{Evaluation Metrics}

For evaluation, we use the area under the
receiver operating characteristic curve (AUROC).
A higher value indicates better performance, with 50\% serving as the baseline.

\vspace{-5pt}
\subsection{Findings on ImageNet-X}

The OOD detection performance (AUROC$\uparrow$) for the conventional ImageNet-1k-based benchmarks and the proposed three benchmarks is shown in \cref{tab:compare_ood_result}. 
For the conventional benchmarks, the ID data consisted of all 1,000 classes from the ImageNet-1k test set. 
Common-OOD data included iNaturalist~\cite{iNaturalist_van2018inaturalist}, Texture~\cite{Texture_cimpoi2014describing}, and OpenImage-O~\cite{OpenImageO_wang2022vim}, while hard-OOD data included SSBHard~\cite{SSBHard_vaze2022open} and NINCO~\cite{NINCO_bitterwolf2023or}. 
The metric was calculated by considering the OOD datasets in the common-OOD and hard-OOD categories as a single OOD dataset.

\tabCombineOODAll

\noindent\textbf{F1. The effectiveness of methods can vary as the benchmarks.}
From \cref{tab:compare_ood_result}, a comparison between the conventional benchmarks and ImageNet-X reveals that the performance rankings changed across the three benchmarks.
No single winner emerges as the best across all distribution shifts. 
Comparing common-OOD and hard-OOD, hard-OOD showed significantly lower overall performance and changes in ranking. 
While zero-shot methods utilizing OOD prompts outperform others in common-OOD, the best performance in hard-OOD was achieved by CLIPN-A, followed by the few-shot method NegPrompt. 
Conversely, in ImageNet-X, NegPrompt outperformed CLIPN-A, unlike in hard-OOD. 
Additionally, LoCoOp outperformed non-OOD prompt methods (MCM, GL-MCM, CoOp) in common-OOD and hard-OOD but showed lower performance compared to these methods in ImageNet-X, indicating notable ranking shifts.

\figImageNetXFullSpectrum

\vspace{-5pt}
\subsection{Findings on ImageNet-FS-X}
\noindent\textbf{F2. The performance drops due to the impact of covariate shifts.}
When comparing ImageNet-X and ImageNet-FS-X in \cref{tab:compare_ood_result}, it is observed that in all methods, the performance significantly drops when covariate shift is introduced. 
These findings indicate that CLIP, even when pre-trained on large datasets, is influenced by covariate shifts, reflecting inherent biases toward specific distributions.
The issue of robustness to covariate shifts in CLIP-based classification has been discussed~\cite{clipood_shu2023clipood}, but it must also be considered in OOD detection.

\noindent\textbf{F3. Few-shot learning induces performance bias due to covariate distribution in training data.}
The performance drop from ImageNet-X to ImageNet-FS-X for each method is shown in \cref{fig:compare_sa_imagenet}. 
From the graph, it can be seen that the differences are larger for few-shot methods compared to zero-shot methods. 
This suggests that training with specific data tends to cause a slight performance bias toward the covariate distribution inherent in the training data.

\noindent\textbf{F4. The method performance ranking remains consistent with constant semantic shift.}
As shown in \cref{tab:compare_ood_result}, no significant changes were observed in performance rankings when comparing ImageNet-X and ImageNet-FS-X.
Calculating the Spearman rank-order correlation coefficient between the benchmarks resulted in a value of 0.90.
This suggests that including both semantic and covariate shifts does not substantially alter the overall ranking of methods compared to evaluations based only on semantic shifts. 
Consequently, while few-shot prompt learning introduces some bias toward the training data's covariate distribution, the overall ranking remained largely unchanged.

\tabWildsAccuracy

\vspace{-7pt}
\subsection{Findings on Wilds-FS-X}\vspace{-5pt}
\noindent\textbf{F5. Wilds is a highly challenging benchmark with low classification and OOD detection performance.}
First, the classification accuracy of the CLIP-based methods in the Wilds-FS-X is shown in \cref{tab:compare_wilds_accuracy}. 
The accuracy was measured on the ID data within the Wilds-FS-X. 
According to the result, while few-shot learning improves classification accuracy, the iWildCam and FMoW datasets remain challenging and insufficiently addressed by current methods. 
Furthermore, as shown in \cref{tab:compare_ood_result}, there is substantial room for improvement in OOD detection performance. 
Particularly for FMoW, the performance remains close to the baseline, indicating significant difficulty in distinguishing between ID and OOD data.
These results indicate that current CLIP-based methods still need improvement in detecting the subtle semantic shifts required for real-world scenarios in the WILDS datasets.

\noindent\textbf{F6. Improved classification accuracy does not necessarily lead to better OOD detection performance.}
From \cref{tab:compare_ood_result} and \cref{tab:compare_wilds_accuracy}, the ranking of OOD detection performance differs from that of classification accuracy.
In particular, for iWildCam, classification accuracy improves significantly with few-shot compared to zero-shot. 
However, in terms of OOD detection, zero-shot methods surpassed few-shot methods in OOD detection, as additional learning improves classification accuracy but hinders detecting subtle semantic shifts between ID and OOD data.
These findings indicate that classification and OOD detection on real-world application data remain challenging and require further improvement. 
Notably, higher classification accuracy does not always mean better OOD detection, emphasizing the need for integrated evaluation of both metrics.

\vspace{-10pt}
\section{Conclusion}\vspace{-7pt}
\label{sec:conclusion}

In this paper, we proposed three challenging and practical OOD detection benchmarks to support a precise understanding of method characteristics and simulations reflecting real-world conditions.
Our experiments on CLIP-based OOD detection methods showed significant performance drops under covariate shifts and real-world-like datasets, revealing room for improvement.
Future work should focus on evaluating methods under more challenging, real-world conditions using our proposed benchmarks to identify limitations and guide improvements, enabling broader applicability of OOD detection methods.

% \vfill\pagebreak


\begin{spacing}{0.9}
\ninept
% \small
% \vspace{-10pt}
% \section{Acknowledgement}
\vspace{-5pt}
\begin{center}
{\bf\normalsize ACKNOWLEDGEMENT\par}
\end{center}
\vskip 0.2em
\vspace{-10pt}

This work was partially supported by JST JPMJCR22U4 and JSPS 25H01164.

\begingroup
\renewcommand{\section}[2]{}  % disable \section temporarily
\begin{center}

\vspace{-10pt}
{\bf\normalsize REFERENCES\par}
\end{center}
% \vskip 0.05em
% \vspace{-5pt}
\bibliographystyle{IEEEbib}
\bibliography{custom}

\begin{thebibliography}{10}

\bibitem{AIkiken_amodei2016concrete}
Dario Amodei, Chris Olah, Jacob Steinhardt, Paul Christiano, John Schulman, and Dan Man{\'e},
\newblock ``Concrete problems in ai safety,''
\newblock {\em arXiv preprint arXiv:1606.06565}, 2016.

\bibitem{AIkiken2_mohseni2022taxonomy}
Sina Mohseni, Haotao Wang, Chaowei Xiao, Zhiding Yu, Zhangyang Wang, and Jay Yadawa,
\newblock ``Taxonomy of machine learning safety: A survey and primer,''
\newblock {\em ACM Computing Surveys}, vol. 55, no. 8, pp. 1--38, 2022.

\bibitem{OODSurvey_yang2024generalized}
Jingkang Yang, Kaiyang Zhou, Yixuan Li, and Ziwei Liu,
\newblock ``Generalized out-of-distribution detection: A survey,''
\newblock {\em IJCV}, vol. 132, no. 12, pp. 5635--5662, 2024.

\bibitem{CLIP_radford2021learning}
Alec Radford, Jong~Wook Kim, Chris Hallacy, Aditya Ramesh, Gabriel Goh, Sandhini Agarwal, Girish Sastry, Amanda Askell, Pamela Mishkin, Jack Clark, Gretchen Krueger, and Ilya Sutskever,
\newblock ``Learning transferable visual models from natural language supervision,''
\newblock in {\em ICML}, 2021.

\bibitem{MCM_ming2022delving}
Yifei Ming, Ziyang Cai, Jiuxiang Gu, Yiyou Sun, Wei Li, and Yixuan Li,
\newblock ``Delving into out-of-distribution detection with vision-language representations,''
\newblock in {\em NeurIPS}, 2022.

\bibitem{LoCoOp_miyai2024locoop}
Atsuyuki Miyai, Qing Yu, Go~Irie, and Kiyoharu Aizawa,
\newblock ``Locoop: Few-shot out-of-distribution detection via prompt learning,''
\newblock in {\em NeurIPS}, 2023.

\bibitem{ImageNet1k_russakovsky2015imagenet}
Olga Russakovsky, Jia Deng, Hao Su, Jonathan Krause, Sanjeev Satheesh, Sean Ma, Zhiheng Huang, Andrej Karpathy, Aditya Khosla, Michael Bernstein, Alexander~C. Berg, and Li~Fei-Fei,
\newblock ``Imagenet large scale visual recognition challenge,''
\newblock {\em IJCV}, vol. 115, pp. 211--252, 2015.

\bibitem{iNaturalist_van2018inaturalist}
Grant Van~Horn, Oisin Mac~Aodha, Yang Song, Yin Cui, Chen Sun, Alex Shepard, Hartwig Adam, Pietro Perona, and Serge Belongie,
\newblock ``The inaturalist species classification and detection dataset,''
\newblock in {\em CVPR}, 2018.

\bibitem{Texture_cimpoi2014describing}
Mircea Cimpoi, Subhransu Maji, Iasonas Kokkinos, Sammy Mohamed, and Andrea Vedaldi,
\newblock ``Describing textures in the wild,''
\newblock in {\em CVPR}, 2014.

\bibitem{GLMCM_miyai2023zero}
Atsuyuki Miyai, Qing Yu, Go~Irie, and Kiyoharu Aizawa,
\newblock ``Gl-mcm: Global and local maximum concept matching for zero-shot out-of-distribution detection,''
\newblock {\em IJCV}, 2025.

\bibitem{NegPrompt_jiangnegative}
Xue Jiang, Feng Liu, Zhen Fang, Hong Chen, Tongliang Liu, Feng Zheng, and Bo~Han,
\newblock ``Negative label guided ood detection with pretrained vision-language models,''
\newblock in {\em ICLR}, 2024.

\bibitem{EOE_caoenvisioning}
Chentao Cao, Zhun Zhong, Zhanke Zhou, Yang Liu, Tongliang Liu, and Bo~Han,
\newblock ``Envisioning outlier exposure by large language models for out-of-distribution detection,''
\newblock in {\em ICML}, 2024.

\bibitem{CLIPN_wang2023clipn}
Hualiang Wang, Yi~Li, Huifeng Yao, and Xiaomeng Li,
\newblock ``Clipn for zero-shot ood detection: Teaching clip to say no,''
\newblock in {\em ICCV}, 2023.

\bibitem{NegativePrompts_li2024learning}
Tianqi Li, Guansong Pang, Xiao Bai, Wenjun Miao, and Jin Zheng,
\newblock ``Learning transferable negative prompts for out-of-distribution detection,''
\newblock in {\em CVPR}, 2024.

\bibitem{IDLike_bai2024id}
Yichen Bai, Zongbo Han, Bing Cao, Xiaoheng Jiang, Qinghua Hu, and Changqing Zhang,
\newblock ``Id-like prompt learning for few-shot out-of-distribution detection,''
\newblock in {\em CVPR}, 2024.

\bibitem{FullSpec_yang2023full}
Jingkang Yang, Kaiyang Zhou, and Ziwei Liu,
\newblock ``Full-spectrum out-of-distribution detection,''
\newblock {\em IJCV}, vol. 131, no. 10, pp. 2607--2622, 2023.

\bibitem{WILDS_wilds2021}
Pang~Wei Koh, Shiori Sagawa, Henrik Marklund, Sang~Michael Xie, Marvin Zhang, Akshay Balsubramani, Weihua Hu, Michihiro Yasunaga, Richard~Lanas Phillips, Irena Gao, Tony Lee, Etienne David, Ian Stavness, Wei Guo, Berton~A. Earnshaw, Imran~S. Haque, Sara Beery, Jure Leskovec, Anshul Kundaje, Emma Pierson, Sergey Levine, Chelsea Finn, and Percy Liang,
\newblock ``{WILDS}: A benchmark of in-the-wild distribution shifts,''
\newblock in {\em ICML}, 2021.

\bibitem{MSP_hendrycks2022baseline}
Dan Hendrycks and Kevin Gimpel,
\newblock ``A baseline for detecting misclassified and out-of-distribution examples in neural networks,''
\newblock in {\em ICLR}, 2017.

\bibitem{mos_huang2021mos}
Rui Huang and Yixuan Li,
\newblock ``Mos: Towards scaling out-of-distribution detection for large semantic space,''
\newblock in {\em CVPR}, 2021.

\bibitem{SSBHard_vaze2022open}
S~Vaze, K~Han, A~Vedaldi, and A~Zisserman,
\newblock ``Open-set recognition: A good closed-set classifier is all you need?,''
\newblock in {\em ICLR}, 2022.

\bibitem{NINCO_bitterwolf2023or}
Julian Bitterwolf, Maximilian Mueller, and Matthias Hein,
\newblock ``In or out? fixing imagenet out-of-distribution detection evaluation,''
\newblock {\em arXiv preprint arXiv:2306.00826}, 2023.

\bibitem{OpenoodV15_zhang2023openood}
Jingyang Zhang, Jingkang Yang, Pengyun Wang, Haoqi Wang, and Yueqian~others Lin,
\newblock ``Openood v1. 5: Enhanced benchmark for out-of-distribution detection,''
\newblock in {\em NeurIPS 2023 Workshop on Distribution Shifts: New Frontiers with Foundation Models}.

\bibitem{ImageNetOOD_yangimagenet}
William Yang, Byron Zhang, and Olga Russakovsky,
\newblock ``Imagenet-ood: Deciphering modern out-of-distribution detection algorithms,''
\newblock in {\em ICLR}, 2024.

\bibitem{opensetImagenet_palechor2023large}
Andres Palechor, Annesha Bhoumik, and Manuel G{\"u}nther,
\newblock ``Large-scale open-set classification protocols for imagenet,''
\newblock in {\em WACV}, 2023.

\bibitem{wordnet}
George~A. Miller,
\newblock ``Wordnet: An electronic lexical database.,''
\newblock {\em MIT press}, 1998.

\bibitem{Imagenetv2_recht2019imagenet}
Benjamin Recht, Rebecca Roelofs, Ludwig Schmidt, and Vaishaal Shankar,
\newblock ``Do imagenet classifiers generalize to imagenet?,''
\newblock in {\em ICML}, 2019.

\bibitem{ImagenetR_hendrycks2021many}
Dan Hendrycks, Steven Basart, Norman Mu, Saurav Kadavath, Frank Wang, Evan Dorundo, Rahul Desai, Tyler Zhu, Samyak Parajuli, Mike Guo, Dawn Song, Jacob Steinhardt, and Justin Gilmer,
\newblock ``The many faces of robustness: A critical analysis of out-of-distribution generalization,''
\newblock in {\em ICCV}, 2021.

\bibitem{ImageNetC_hendrycks2019robustness}
Dan Hendrycks and Thomas Dietterich,
\newblock ``Benchmarking neural network robustness to common corruptions and perturbations,''
\newblock in {\em ICLR}, 2018.

\bibitem{covariateParadox_long2024rethinking}
Xingming Long, Jie Zhang, Shiguang Shan, and Xilin Chen,
\newblock ``Rethinking the evaluation of out-of-distribution detection: A sorites paradox,''
\newblock {\em arXiv preprint arXiv:2406.09867}, 2024.

\bibitem{wildcapture_10350906}
Luca Cultrera, Lorenzo Seidenari, and Alberto Del~Bimbo,
\newblock ``Leveraging visual attention for out-of-distribution detection,''
\newblock in {\em ICCVW}, 2023.

\bibitem{OODVlmSurvey_miyai2024generalized}
Atsuyuki Miyai, Jingkang Yang, Jingyang Zhang, Yifei Ming, Yueqian Lin, Qing Yu, Go~Irie, Shafiq Joty, Yixuan Li, Hai Li, Ziwei Liu, Toshihiko Yamasaki, and Kiyoharu Aizawa,
\newblock ``Generalized out-of-distribution detection and beyond in vision language model era: A survey,''
\newblock {\em arXiv preprint arXiv:2407.21794}, 2024.

\bibitem{CoOp_zhou2022learning}
Kaiyang Zhou, Jingkang Yang, Chen~Change Loy, and Ziwei Liu,
\newblock ``Learning to prompt for vision-language models,''
\newblock {\em IJCV}, vol. 130, no. 9, pp. 2337--2348, 2022.

\bibitem{OpenImageO_wang2022vim}
Haoqi Wang, Zhizhong Li, Litong Feng, and Wayne Zhang,
\newblock ``Vim: Out-of-distribution with virtual-logit matching,''
\newblock in {\em CVPR}, 2022.

\bibitem{clipood_shu2023clipood}
Yang Shu, Xingzhuo Guo, Jialong Wu, Ximei Wang, Jianmin Wang, and Mingsheng Long,
\newblock ``Clipood: Generalizing clip to out-of-distributions,''
\newblock in {\em ICML}, 2023.

\end{thebibliography}
\endgroup
\end{spacing}

\normalsize
\beginsupplement

\appendix
\section*{Appendix}

\section{Experimental Detail}

In this section, we describe the details of each CLIP-based out-of-distribution (OOD) detection method employed in our experiments.

\subsection{Zero-shot OOD detection}

\noindent\textbf{MCM and GL-MCM.}
We utilized the pre-trained CLIP-ViT-B/16 model (\url{https://github.com/openai/CLIP}).
The same pre-trained CLIP model was consistently used across the other methods as well.

\noindent\textbf{NegLabel.}
For the list of additional labels, we adopted the label set provided by previous studies.

\noindent\textbf{EOE.}
This method uses a large language model (LLM) to generate additional labels.
We employed gpt-3.5-turbo-16k for this purpose.
Following prior work, which suggests designing LLM prompts based on visual similarity to produce suitable outlier labels, we used two types of prompts: the \textit{Far OOD prompt}, which generates domain-level labels for Common-OOD evaluation; and the \textit{Near OOD prompt}, which generates fine-grained labels for Hard-OOD and our proposed benchmark.

\noindent\textbf{CLIPN.}
We employed the ``no'' Text Encoder provided by previous work as a pre-trained model.
Among the two proposed OOD scoring algorithms—the \textit{Competing-to-win Algorithm} and the \textit{Agreeing-to-differ Algorithm}—we chose the latter (CLIPN-A), as it has been reported to deliver superior detection performance.

\subsection{Few-shot OOD detection}
The results for each few-shot method were averaged over 3 random seeds.

\noindent\textbf{CoOp and LoCoOp.}
We used the pre-trained CLIP-ViT-B/16 model and trained with 16 shots, following the same hyperparameters as in prior work (\eg training epochs = 50, learning rate = 0.002, batch size = 32, token length = 16).

\noindent\textbf{NegPrompt.}
Prompt learning was conducted in two stages: first, positive prompts were trained for 100 epochs; then, with the positive prompts frozen, negative prompts were trained for 10 epochs.
The loss weights ($\beta$ = 0.1, $\gamma$ = 0.05) were taken from previous work.
Training was performed with 16 shots. Since the learning rate and batch size were not explicitly stated, we set them to 0.002 and 32, respectively.

\noindent\textbf{IDPrompt.}
Due to its higher training cost, training was conducted with only 1 shot, as in prior studies.
Each image was randomly cropped into 256 patterns, and the top 32 were used for training.
Other hyperparameters (\eg training epochs = 3, learning rate = 0.005, batch size = 1) also followed previous implementations.

\section{OOD Score examples}

In this section, we present the out-of-distribution (OOD) score distributions for the benchmarks along with several example samples.

\figImageNetXWithSample
\subsection{ImageNet-X}

\cref{fig:distribution_imagenet_x_ver2} presents the OOD score distributions and sample scores in ImageNet-X.
The distributions of MCM, CoOp, and NegPrompt are shown as representative examples of CLIP-based methods.
Furthermore, one representative sample from both the ID and OOD data is provided as an example.

A comparison of the distributions and sample scores reveals that detection performance varies significantly across methods, even when applied to the same data.
For example, with NegPrompt, both samples are correctly identified.
However, with MCM and CoOp, distinguishing between the two images as ID and OOD becomes challenging.
Notably, CoOp assigns a lower OOD score to (i) than to (ii), resulting in reversed detection outcomes compared to NegPrompt, highlighting substantial differences in performance between the methods.

\subsection{ImageNet-FS-X}
\figImageNetFSXWithSample

\cref{fig:distribution_imagenet_fs_x_ver2} shows the OOD score distribution and sample score examples in ImageNet-FS-X.
The samples are drawn from ImageNet-FS-X, consisting of (i) ImageNet-1k, (ii) ImageNet-V2, (iii) ImageNet-R, and (iv) ImageNet-C, with examples selected based on identical labels across these datasets.
These samples belong to the ``piano'' label and are part of the ID data.

From the sample scores, covariate-shifted ID data is shown to be detectable to some extent, even relative to Training ID. 
Notably, NegPrompt, which performed best on ImageNet-FS-X, consistently detects all samples as ID with high confidence.
Conversely, methods like MCM and CoOp struggle to classify (ii), (iii), and (iv) as ID, though (i) is accurately detected. 
Similar to ImageNet-X, detection performance varies by method and image. 
Additionally, while MCM and NegPrompt show scores decreasing in the order of (i), (ii), (iii), and (iv), CoOp assigns a higher score to (iii) than (ii), highlighting differences in recognition order among methods.

Next, we examine the impact of covariate shifts. 
First, in terms of score distribution, the distributions of ID and OOD in ImageNet-FS-X are closer compared to ImageNet-X.
Specifically, the ID data distribution exhibits greater variance, suggesting that covariate shifts cause some ID samples to have scores closer to OOD data. 
Additionally, in the sample examples, (iv) is an image with noise added to (i). 
Comparing (i) and (iv), we see that (iv) has a significantly lower score than (i), making detection more difficult. 
These results demonstrate that differences in style and quality substantially affect OOD detection performance.

\section{Covariate shifts analysis}

In this section, we compare the OOD detection performance of the source datasets in the full-spectrum benchmarks.
This analysis enables further insights into performance variations caused by the covariate shifts and the biases introduced by few-shot learning.

\figImageNetFSCompare
\subsection{ImageNet-FS-X}

\cref{fig:compare_imagenet_fs} presents the OOD detection performance across source datasets in ImageNet-FS-X.
Here, among the variant datasets of ImageNet, ImageNet-V2 represents resampling bias, ImageNet-R reflects style changes, and ImageNet-C serves as an example of image corruption. 
In all methods, ImageNet-1k achieves the highest performance, followed by ImageNet-V2, ImageNet-R and ImageNet-C.
Notably, the AUROC for ImageNet-C is more than 20\% lower than ImageNet-1k.
These results indicate that CLIP-based methods are sensitive to changes in the covariate distribution of the data, resulting in significant performance variations.

Next, focusing on the relationship between method types and performance variations, zero-shot methods show a consistent trend of performance changes: ImageNet-1k\(>\)ImageNet-V2\(\fallingdotseq \)ImageNet-R\(>\)ImageNet-C.
This suggests that the pre-trained weights of CLIP inherently favor this order of performance across datasets.
In contrast, few-shot methods exhibit a greater performance gap between ImageNet-V2 and ImageNet-R, which was less pronounced in zero-shot methods.
Specifically, CoOp and NegPrompt demonstrate a noticeable drop in performance on ImageNet-R compared to ImageNet-V2.
While ImageNet-V2 consists of natural images similar to ImageNet-1k, ImageNet-R includes objects represented in alternative styles, such as illustrations and sculptures, resulting in significant distributional differences.
These results indicate that few-shot training amplifies the dependence on the covariate distribution of the training dataset, making the methods more susceptible to covariate shifts.

\figWildsFSCompare
\subsection{Wilds-FS-X}

The OOD detection performance for each dataset group, \textit{Test} and \textit{Test-ID}, in Wilds-FS-X is shown in \cref{fig:compare_wilds_fs}.
First, in zero-shot methods, all datasets demonstrate better performance on the covariate-shifted ID data, \textit{Test}, compared to the training ID data, \textit{Test-ID}.
This result indicates that the CLIP weights are inherently biased toward achieving higher performance on \textit{Test}.
Furthermore, the observed performance variation due to covariate shifts highlights the need to consider robustness against covariate shifts that can occur in real-world scenarios.

On the other hand, in the case of few-shot methods, \textit{Test} did not always exhibit higher performance than \textit{Test-ID}.
Notably, in iWildCam, methods such as CoOp, LoCoOp, and IDPrompt achieved higher performance on \textit{Test-ID} compared to \textit{Test}.
These results further confirm that, even in Wilds-FS-X, few-shot methods exhibit a performance bias toward the covariate distribution of the training ID data.

From the analysis, CLIP's pre-trained knowledge is biased toward specific covariate distributions, affecting OOD detection performance under covariate shifts.
Furthermore, across all benchmarks, a performance bias toward the covariate distribution of the training data was observed when using few-shot learning.
Future discussions on the robustness of CLIP-based OOD detection methods to covariate shifts should consider both the biases in pre-trained models and the effects of additional training.

\end{document}